\documentclass[conference]{IEEEconf}

\input epsf
\usepackage{graphicx}
\usepackage{subfigure}
\usepackage{cite}
\usepackage{amsmath,amssymb,amsfonts}
\usepackage{algorithmic}
\usepackage{textcomp}
\usepackage{xcolor}
\usepackage{color}
\usepackage{siunitx}
\usepackage{multirow}
\usepackage{booktabs}
\usepackage{float}
\usepackage{cleveref}
\usepackage{algorithm2e} 
\usepackage{url}

\begin{document}
\IEEEoverridecommandlockouts

\title{\textbf{\Large AuthROS: Secure Data Sharing Among Robot Operating Systems Based on Ethereum\\}}

\author{Shenhui Zhang$^{1}$, Wenkai Li$^{1}$, Xiaoqi Li$^{1,*}$, Boyi Liu$^{2,*}$\\
	\normalsize $^{1}$Key Laboratory of Internet Information Retrieval of Hainan Province, Hainan University, Haikou, China\\
	\normalsize $^{2}$Robotics Institute, The Hong Kong University of Science and Technology, Hong Kong SAR, China\\
	\normalsize csxqli@gmail.com, by.liu@ieee.org\\
	\thanks {*corresponding author}
}



\maketitle
\begin{abstract}
The Robot Operating System (ROS) streamlines human processes, increasing the efficiency of various production tasks. However, the security of data transfer operations in ROS is still in its immaturity. Securing data exchange between several robots is a significant problem. This paper proposes \textit{AuthROS}, an Ethereum blockchain-based secure data sharing method, for robot communication. It is a ROS node authorization system capable of ensuring the immutability and security of private data flow between ROS nodes of any size. To ensure data security, AuthROS employs the smart contract for permission granting and identification, SM2-based key exchange, and SM4-based plaintext encryption techniques. In addition, we deploy a data digest upload technique to optimize data query and upload performance. Finally, the experimental findings reveal that AuthROS has strong security, time performance, and node forging in cases where data should be recorded and robots need to remain immobile.

\end{abstract}


\IEEEpeerreviewmaketitle

\section{Introduction}
The Robot Operating System (ROS)~\cite{ref1} is an open-source meta-operating system for robots. It is a distributed multi-process framework based on message communication. ROS is designed to improve the code reuse rate in the field of robotics research and development. ROS provides functions similar to those provided by operating systems (OS), such as hardware abstraction description, low-level driver management, etc. The essence of ROS is a TCP/IP-based Socket communication mechanism~\cite{ref27}, which is capable of performing several types of communication, such as service-based synchronous RPC communication, topic-based asynchronous data stream communication, and parameter server-based data storage. This flexible framework enables different modules of ROS to be designed separately and loosely coupled at runtime.

However, there are some obvious drawbacks to ROS~\cite{ref3,ref4,ref5}. First of all, ROS lacks measures to ensure data security. Attackers can steal data from ROS utilizing the Publisher-Subscriber mode, which endangers the immobility and safety of data. In addition, ROS also has problems with data reliability, and data passed among ROS nodes based on this framework can be intercepted or forged~\cite{ref25}. For example, these defects in autonomous driving would result in a serious accident. These two defects lead to insecure and unstable data exchange and sharing in the scenario of multi-robot interaction based on ROS.

Furthermore, the issues raised by ROS had been alleviated with the assistance of SROS~\cite{SROS}, a typical scheme based on Transport Layer Security (TLS) and certificate mechanisms. In the ROS 2.0 phase, integrating Data Distribution Service (DDS) components and SROS enhances the scheme's authorization and access control security. ROS 2.0 implements centralized access control utilizing policies including permission control and certificate signing. The DDS security standard indicators~\cite{DDS_security} include Authentication, Access Control, Cryptographic, Logging, and Data Tagging. The ROS 2.0 have not achieved Logging and Data Tagging, specifically, the capacity to record data and behaviour. As a result, when an attacker gains central control, the data can be manipulated directly, and there is no way to trace back error information.

\begin{figure}[t]
\setlength{\abovecaptionskip}{0cm} 
\setlength{\belowcaptionskip}{-5cm}
    \centering
    \includegraphics[width=0.5\textwidth]{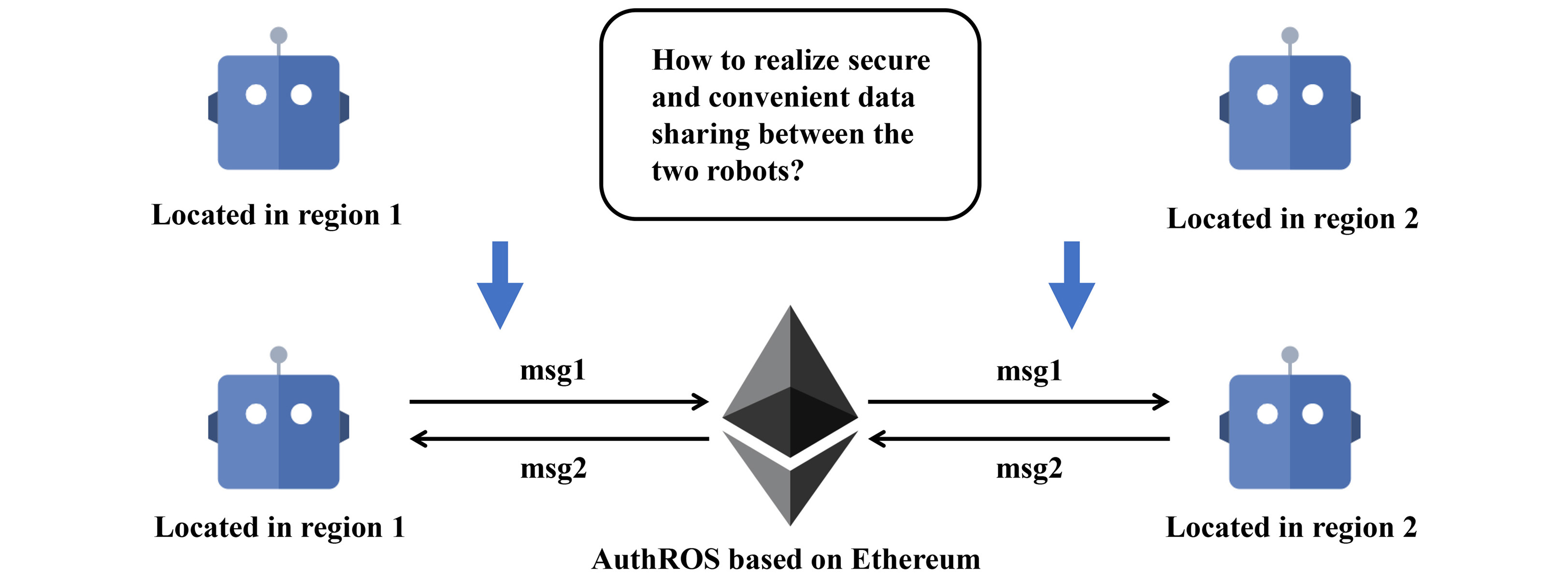}
    \caption{The design idea of the AuthROS framework. To reduce the risk brought by attacks during data sharing in ROS, Ethereum is used to design this data sharing framework.}
    \vspace{-3.5ex}
    \label{fig:fig1}
\end{figure}

Therefore, we leverage Ethereum blockchain~\cite{ref6} to solve the above drawbacks and propose a novel framework named AuthROS (Authority in Robot Operating Systems), whose design idea is shown in Fig. \ref{fig:fig1}. In order to ensure the security of the entire communication network and data, AuthROS leverages a series of remarkable cryptographic algorithms, i.e., the SM algorithm family~\cite{SM_algorithms}. At present, three types of algorithms are mainly applied: SM2, SM3, and SM4~\cite{SM_algorithms}. Based on the SM algorithms and blockchain technology, we can finally solve the existing defects of ROS. AuthROS achieves an efficient and secure data sharing framework for ROS based on blockchain technology, Web3, and SM algorithms. And due to the data traceability of blockchain systems, data logging for monitoring abnormal node behavior can be achieved. It allows critical data captured by robots shared by other robots with authority in the same Ethereum network.

At the same time, the confidential data is also stored in the Ethereum network for later check. Furthermore, it has virtually no restrictions on the types of data, most types of messages are supported to interact with the Ethereum network. The algorithm encryption system based on the SM algorithm family can effectively ensure the security of data transmission.

The main contributions of this paper are as follows:
\begin{itemize}
    \item [(1)] We propose a novel framework called AuthROS based on blockchain technology. To the best of our knowledge, it is the \textit{first} secure data-sharing framework for robots loaded with ROS (Section IV).
    \item [(2)] AuthROS has a functionality of authority granting controlling access to specified confidential data transmitted among ROS. Furthermore, it can conduct secure encrypted communication leveraging SM algorithms to prevent attacks, such as Node Forging (Section V-C).
    \item [(3)] AuthROS achieves a secure, reliable, and convenient interaction solution for ROS-based robots leveraging the Ethereum blockchain. Evaluation of the process for generating digest from 800KB encrypted data reveals that AuthROS is efficient, completing in 6.34ms (Section V).
\end{itemize}

\section{Background}
In this section, we introduce the ROS system and the Ethereum network, since AuthROS consists of them. 
\label{sec:background}

\subsection{ROS.}
ROS is a distributed process framework to promote the high reusability of robotic software systems. It is an open-source, meta-operating system that provides adaptable and practical qualities for robot manipulations~\cite{ROS2}. 
Some abstractions in ROS developed from the OS offer services comparable to the operating system, including standard hardware APIs, low-level device management, message transmission between nodes, and package management for application distribution. 
ROS also includes a Peer-to-Peer (P2P) network topology that blends service-based synchronous Remote Procedure Call (RPC) communication, topic-based asynchronous data flow communication, and others.

\textbf{Terms.}
Each of the software modules is a node~\cite{ref1}. And the nodes communicate with each other by passing messages, which are strongly typed and support multiple nesting.
Another "odometry" or "map" type term is "topic," which refers to a way of communication in nodes, from which numerous publishers and subscribers complete the message transmission.
Finally, a service consists of a string name, a request message, and a response message. However, the network communication protocol it uses does not handle synchronous transactions.

\subsection{Ethereum.}
Ethereum~\cite{ref6}, a popular blockchain platform, is another helpful technology. Blockchain technology aims to record all transactions in the network to safeguard data. It generates users' addresses using elliptic curve algorithms and hashing algorithms before authenticating transactions.

\textbf{Geth.}
Geth is an Ethereum client built in Golang language, and the local machine can join the Ethereum P2P network as a node after the running of Geth~\cite{stan}. In this paper, the Geth is used to build an Ethereum private network. Ethereum supports Externally Owned Accounts (EOA) and smart contract accounts. With the exception of the network administrator, who has a contract account, all AuthROS robots are associated with EOA accounts. The first 20 bytes of the SHA3 hash of a user's public key serve as the account's index~\cite{ref7}.

\textbf{Consensus Algorithms.}
Consensus algorithms provide the immutability, automation, and anonymity of blockchain transactions. Consensus algorithms maintain the meaning and value of blockchain technology as a distributed database. It ensures that the states of the blocks on the chain remain consistent. Proof of Work (PoW)~\cite{defiSecurity, defisecurity_spb}, Proof of Authority (PoA)~\cite{ref8} are the consensus methods for Ethereum in AuthROS. Nodes in networks using PoW and PoA consensus algorithms have different roles as miners or validators. PoW relies on mining operations to validate blocks, whereas PoA employs trusted nodes that are pre-authorized.


\textbf{Smart Contract.}
The smart contract is an executable software program that can be interacted with peers on the network~\cite{survey_of_blockchain2020li,Gaschecker}. It has increased the scalability of blockchain. Users can execute customized transaction rules in smart contracts, and transactions are irreversible once completed~\cite{erasable_accounts}. Additionally, smart contracts can be programmed in a Turing complete language known as Solidity, Vyper, etc~\cite{clue}. Peer-based decision-making is enabled through carefully built smart contracts in applications such as IoT, multi-robot systems, and smart cities.

\section{Related Work}
Large-scale applications of robots will inevitably involve many problems, such as the security of data transmission, data sharing, and the classical Byzantine problem~\cite{ref11}. With the development of ROS 2.0, several studies have concentrated on providing powerful tools for secure robots interactions, and pursuing complete DDS standard indicators~\cite{DDS_security}.

Sundaresan et al.~\cite{Secure_ROS} proposed an access control strategy based on IPSec to solve the problem of identity authentication and encrypted robot communication. It ensured that IP packets were encrypted and authenticated by modifying the master node and client libraries. Nonetheless, it restrictd access due to the control of user permissions through IP. 

Ruffin et al.~\cite{SROS} proposed SROS, which was based on Transport Layer Security (TLS) and certificate authentication mechanisms to achieve identity authentication, encryption, and access control of communication. Thus, ports are assigned to robots at runtime. Multiple robots can access them simultaneously, with centralized access control ensured by the security protocol and identity authentication mechanism.

Combining Datagram TLS and TLS, Breiling et al.~\cite{syscon} proposes a secure channel for node-to-node communication. It requests the initial handshake using the certificate and RSA encryption, which is encrypted using the AES algorithm. And it utilizes Message Authentication Codes (MACs) following data transmission to ensure data integrity.

However, none of the above work has solved the problem of data traceability. Due to the information record function and security performance of blockchain, it has been favored by researchers in the robot community, and much research on the integration of robots and blockchain has been carried out.
 
Some research focusing on swarm communication is listed as follows. To combat COVID-19 and break through the bottleneck of existing multi-swarming UAVs based on 5G, Rajesh Gupta et al.~\cite{ref12} proposed a blockchain-envisioned software multi-swarming UAV communication scheme based on a 6G network with intelligent connectivity. Pranav K. Singh et al.~\cite{ref13} proposed an efficient communication framework for swarm robotics based on PoA consensus to break through the limitations of existing robotic control and communication schemes. Eduardo Castelló Ferrer et al.~\cite{ref14} introduced the first learning framework for secure, decentralized, and computationally efficient data and model sharing among multiple robot units installed at multiple sites. Pramod et al.~\cite{ref15} used a set of experiments to validate that Ethereum can be a secure media for communication for multiple small Unmanned Aerial Vehicles (sUAVs).

Research focus on secure information sharing also weighs a lot. Alsamhi et al.~\cite{ref16} proposed a framework to facilitate information sharing within multi-robot using Ethereum. This framework proved to be effective. Nishida et al.~\cite{ref17} introduced a methodology to share information among autonomous robots and demonstrated through experiments how the differences in data size recorded in the blockchain affect the chain size. Jorge Peña Queralta et al.~\cite{ref18} presented a novel approach to managing collaboration terms in heterogeneous multi-robot systems with blockchain. This approach can estimate the available computational resources of different robots and integrate information about the environment from different robots, to evaluate and rank the quality and accuracy of each of the robots' sensor data. Vasco Lopes et al.~\cite{ref19} proposed an architecture that uses blockchain as a ledger and smart contract for robotic control by using external parties, Oracles, to process data. The proposed architecture shows great potential for secure information sharing between robots.

There comes the classic Byzantine problem with Robotic swarms. In the survey of Eduardo Castelló Ferrer et al.~\cite{ref20}, a set of Byzantine Follow The Leader (BFTL) problems were presented, and algorithms to tackle the BFTL problems based on blockchain were proposed too. Alexandre Pacheco et al.~\cite{ref21} presented a robot swarm composed of Pi-puck robots that maintain a blockchain network. The blockchain served as a security layer to neutralize Byzantine robots. Volker Strobel et al.~\cite{ref22} demonstrated how robotic swarms achieve consensus in the presence of Byzantine robots exploiting blockchain technology.

\section{Framework and Elements of AuthROS}
In this section, we explain the AuthROS framework sharing data based on the blockchain, the SM algorithms, and ROS. AuthROS leverages encryption technology, consensus mechanisms, and smart contract to assure security in the data generation, transmission, and sharing process.

\begin{figure*}[t]
\setlength{\abovecaptionskip}{0cm} 
\setlength{\belowcaptionskip}{-5cm}
    \centering
    \includegraphics[width=0.65\textwidth]{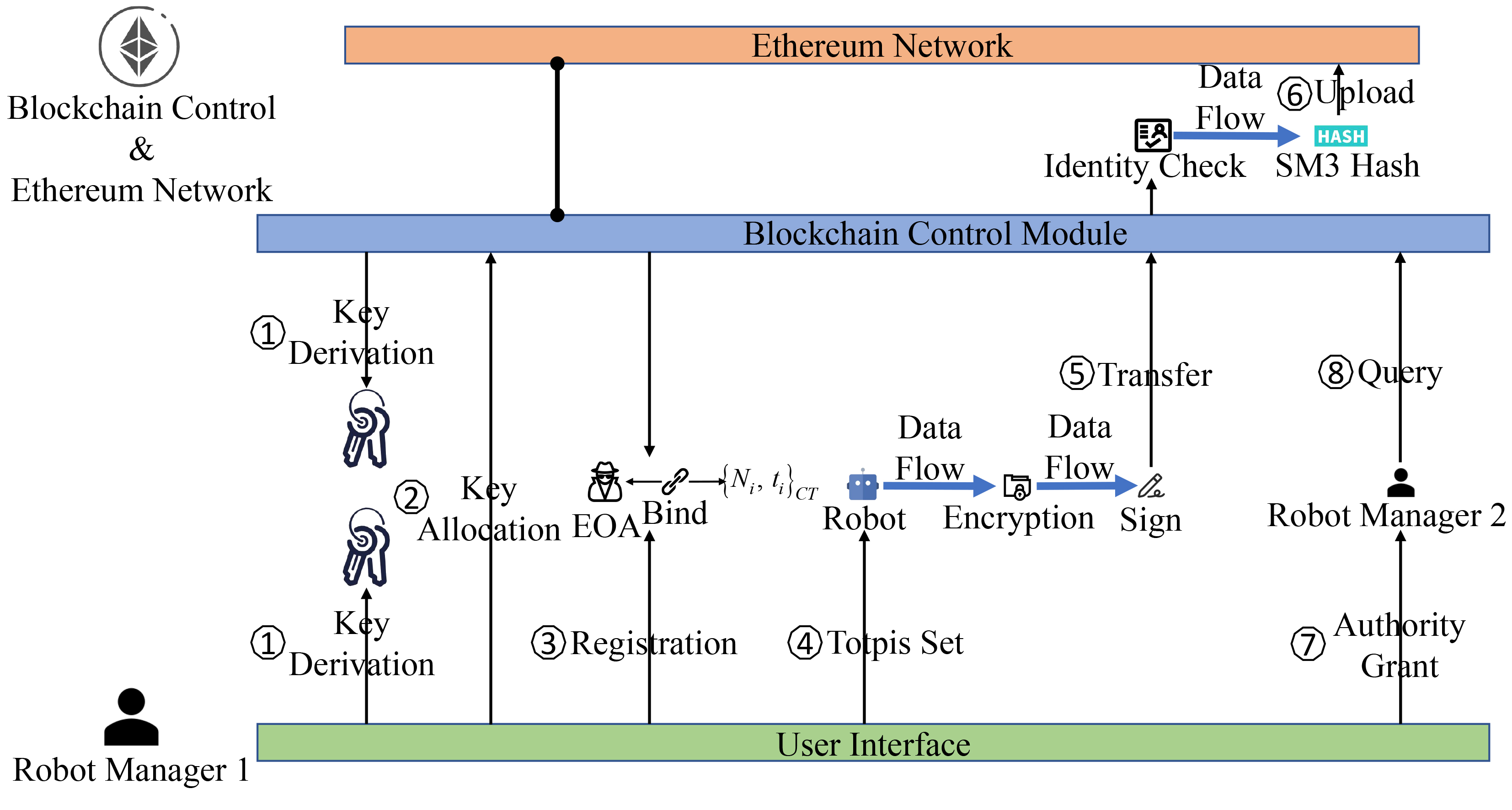}
    \caption{The framework of AuthROS. The User Interface, Blockchain Control Module, and Ethereum Network are represented by the green, blue, and orange boxes, respectively. And the blue and black arrows are the flow of data and operations, respectively.}
    \vspace{-2ex}
    \label{fig:fig2}
\end{figure*}

\subsection{Assumptions.}
To ensure the availability and efficiency of AuthROS, we incorporate some assumptions into its design. The role of AuthROS will likewise be heavily influenced by these assumptions. The following assumptions are made:

    \textbf{Blockchain Security.} Due to the features of distribution and encryption in blockchain, it possesses superior security performance. Therefore, we assume that Ethereum, as a channel for sharing information, is secure and trustworthy.
    
    \textbf{Robots Manager.} We presume that any robot or robot cluster is  capable of maintaining by more than one manager for data communication. These managers refer to the Core Users of AuthROS (CURA), which serves as data sharing.
    
    \textbf{Identity Knowability.} Different CURAs only employ AuthROS for data sharing after establishing a trusting connection. In other words, the foundation of data exchange is that the CURAs are confident with each other. Any CURA cannot reveal the EOA address that confirms its identification to a third party.
    
    \textbf{Unique Means of Sharing.} CURAs will only exchange data via AuthROS. It is essential for the accessibility of AuthROS.
    
    \textbf{Administrator.} There is always an administrator on the Ethereum private network who handles membership additions and other emergencies.

Due to the immutability, semi-decentralization, and anonymity of the private chain, it is ideally suited as a platform for information sharing in AuthROS. The blockchain network and hardware platform will then be described in depth.

\subsection{Blockchain Network.} We select PoA and PoW as the consensus algorithms for AuthROS on Ethereum. The characteristics of blockchain networks based on two distinct consensus algorithms are as follows:

\textbf{Same Contract.} No matter which consensus mechanism the network is employed, we all utilize the same contract for consistent internal interfaces.

\textbf{Same Block Difficulty and Quantity of Users.} To examine the applicability of the consensus technique in the subsequent tests, we not only simulate the same number of users in the blockchain network based on the two consensuses, but also we set the block difficulty to the same value in the genesis block.

Ethereum is utilized to build the blockchain network of AuthROS. The private network of AuthROS is developed based on Geth. The smart contract is deployed with EVM (Ethereum Virtual Machine). PoW and PoA are two different consensus algorithms, and we evaluate the effects of the two on the time performance of AuthROS.
Furthermore, to address the limited computing resources in smart terminal devices, we build a private Ethereum network in the cloud server. The robots can obtain the connection to the blockchain of AuthROS by the local ROS master running on the host with access to the internet.

\begin{figure}[htbp]
\setlength{\abovecaptionskip}{0cm} 
\setlength{\belowcaptionskip}{-5cm}
    \centering
    \includegraphics[width=0.5\textwidth]{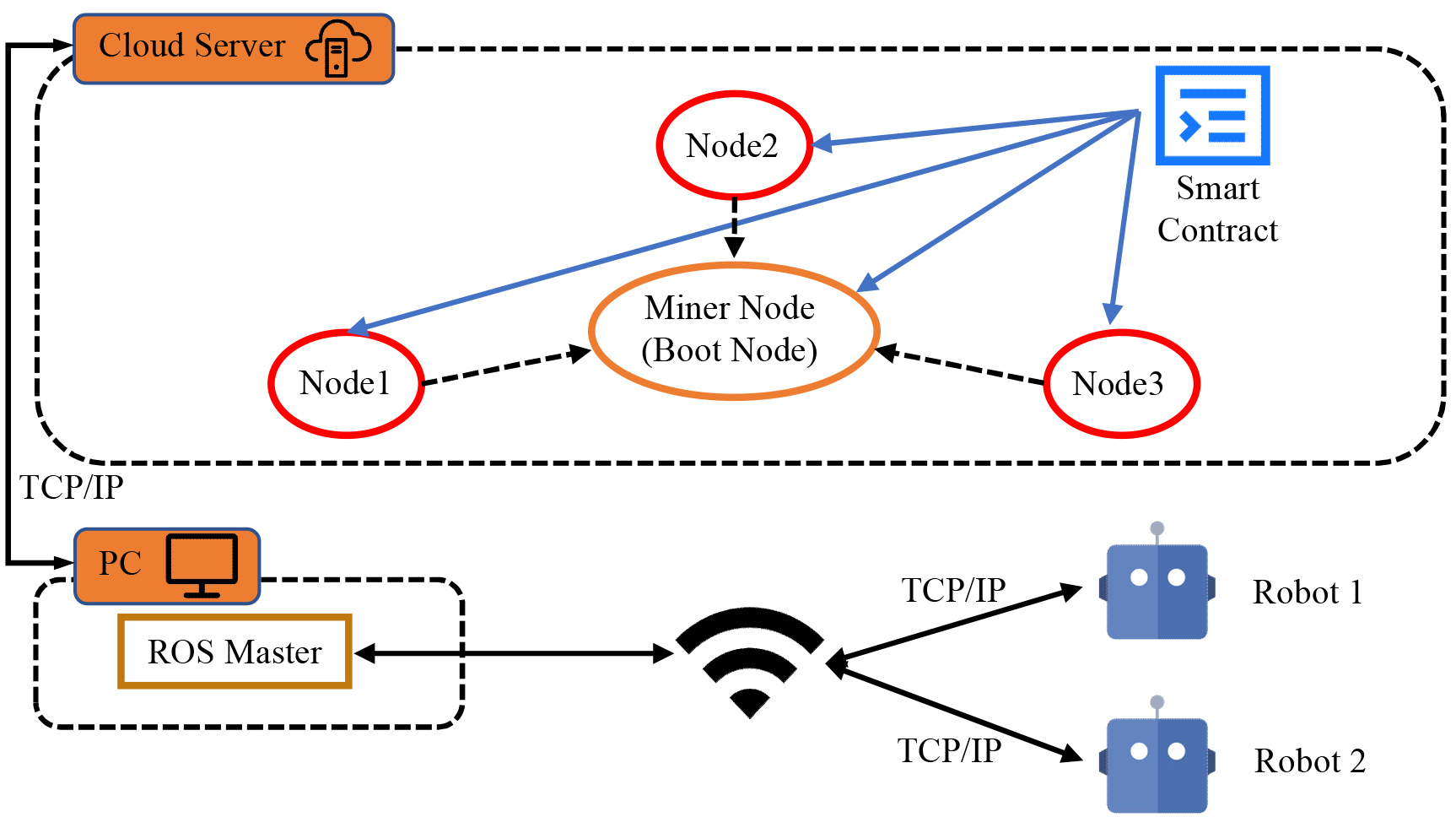}
    \caption{The framework of the private Ethereum network of AuthROS. The blockchain network is deployed in the cloud server and contains three nodes. Miner Node is the boot node, which connects other nodes, such as Node1 and Node2. All robots connect to the blockchain network through a node within the ROS master in local ROS.}
    \vspace{-2ex}
    \label{fig:fig3}
\end{figure}

Due to the robot's limited processing power, building an Ethereum node there will encounter the computational bottleneck typical of Edge Computing. Therefore, as Fig. \ref{fig:fig3} depicted, the Ethereum network composed of 3 nodes is implemented using a host. The sole responsibility of the robot is to maintain contact with the host's ROS Master. If ROS is kept isolated from the Ethereum network, the Edge Computing bottleneck can be ignored. The miner node serves as the bootstrap node and is in charge of bringing together other nodes to create the overlay network. Moreover, to interact with the Ethereum network, four ROS-Melodic robots are connected to two EOA accounts in Node2 and Node3, respectively.

\subsection{AuthROS Framework and Process}
For the process of AuthROS in Fig. \ref{fig:fig2}, users should first upload their name/token key-value pairs. And an SM4 key and an SM2 public key are required to authenticate the digital signature, which is necessary for the Identity Check and Authority Grant. Users can choose a topic to monitor after registering the identification. The monitored topic often forwards some essential information, such as the data in radar, camera, and other sensors. As soon as the topic publishes data, AuthROS will immediately launch a subscriber to capture and parse the topic's contents. The data will be delivered to the network management module of AuthROS after being encrypted by SM4 and signed by SM2. After the validation of data ciphertext by the SM2 signature, the SM3 hash method generates the data digest. The value of the digest will be posted to the blockchain network. Users can grant access to their shared data to other users on the chain. The user's identity is represented by a unique Ethereum external account (EOA) in the Ethereum network, and authority is granted primarily via the exchange of SM4 keys uploaded by the user.

This framework possesses the following characteristics:

    \textbf{Plasticity.} The AuthROS uses a private chain, which is more flexible in terms of block time and consensus conditions. And the semi-decentralized structure of the private chain makes it add new members to the network more conveniently.
    
    \textbf{Process Security.} The AuthROS uses SM4, a symmetric encryption technique, to encrypt all data transmission and interaction operations. The data digest interaction system combined with Identity Check and Authority Grant mechanism can ensure data integrity, security, and immutability.

\begin{figure*}[ht]
\setlength{\abovecaptionskip}{0cm} 
\setlength{\belowcaptionskip}{-5cm}
    \centering
    \includegraphics[width=0.7\textwidth]{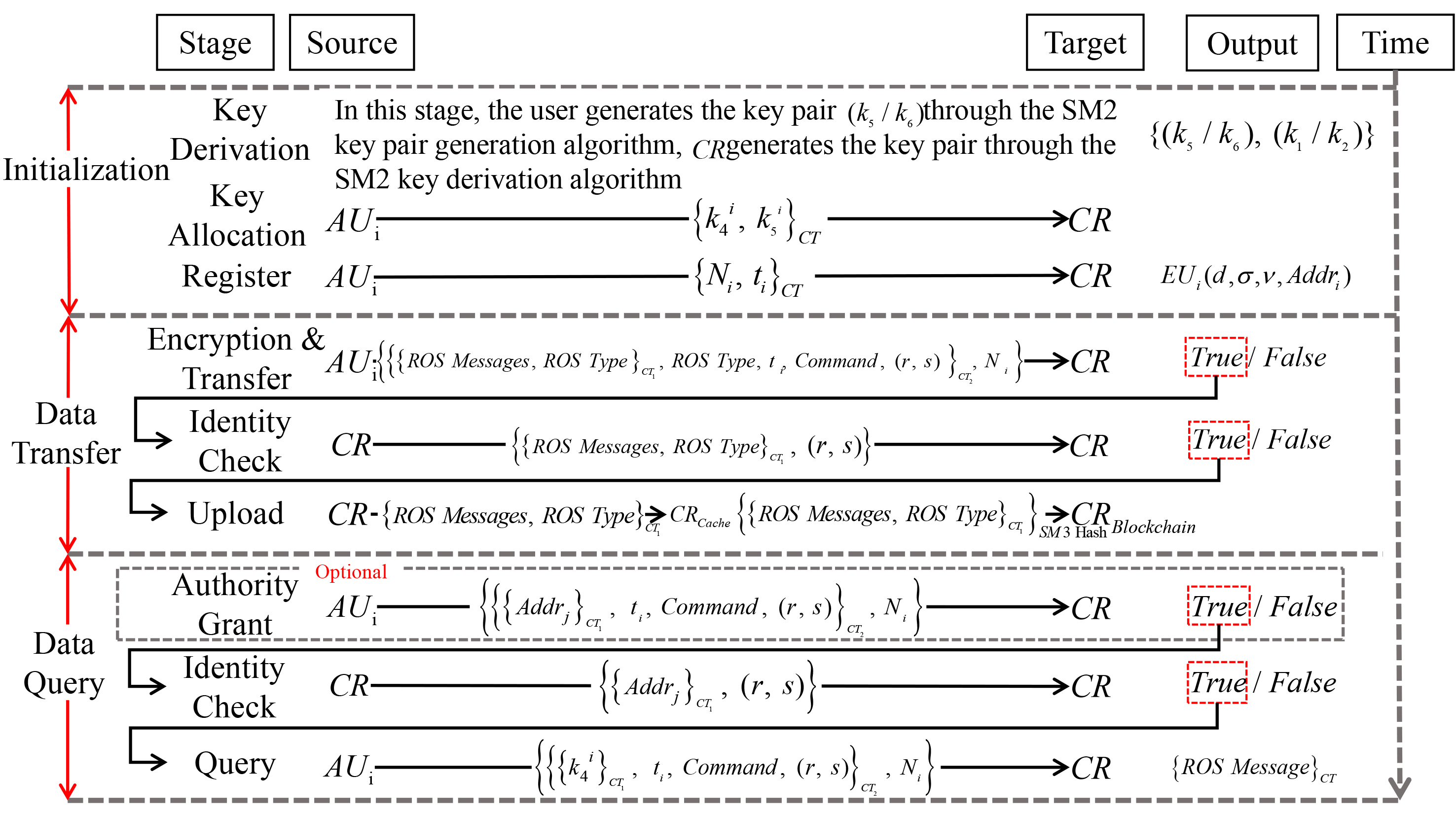}
    \caption{The process of data sharing in AuthROS. The whole process is worked from top to bottom, and each dotted box is a phase.}
    \vspace{-2ex}
    \label{fig:fig6}
\end{figure*}

\subsection{Data Sharing Protocol}
In this section, we will introduce the core mechanism in AuthROS, including the key distribution protocol, data encryption scheme, etc. Due to the possibility of data being intercepted and altered during transmission, there are some stricter standards for data integrity and security in AuthROS. Consequently, AuthROS developed an interaction strategy based on data digests that uses the SM2 digital signature algorithm and the SM3 hash algorithm. In this strategy, the server locally maintains all of the data while the blockchain network uploads the data digest. The notations used in the design are summarized in Table \ref{tab:tab1}.

The design of AuthROS consists of three phases: (1) Initialization, (2) Data Transfer, and (3) Data Query. Fig. \ref{fig:fig6} depicts the subprocess corresponding to each level.
\begin{table*}[htbp]
\setlength{\abovecaptionskip}{0cm} 
\setlength{\belowcaptionskip}{-5cm}
    \centering
    \caption{Summary of notations in this paper.}
    \small
    \begin{tabular}{|l|l|}
        \hline
         \textbf{Notation} & \textbf{Meaning} \\
        \hline
         $AU_{i}$ & The $i_{th}$ user of AuthROS \\
         $Type$ & Data type of ROS topic transmission\\
         $LV_{x, y, z}$ & Three axis velocity of data of odometry\\
         $A V_{x, y, z}$ & Triaxial angular velocity of odometry\\
         $P o s e$ & Pose information contained in data of odometry\\ 
         $C o v$ & Covariance information contained in data of odometry\\
         $T S$ & Timestamp information contained in odometry\\
         $T$ & Time of ROS data captured \\
         $C R$ & Cloud server of AuthROS deploying Ethereum \\
         $N D$ & Plaintext data \\
         $C T$ & Ciphertext of plaintext \\
         $A d d r_{i}$ & Address of the identity of the $i_{th}$ user\\
         $d, \sigma, v$ & Structure for users to store shared data, their own\\ & keys, and keys of other users\\
         $E U_{i}\left(d, \sigma, v, A d d r_{i}\right)$ & Corresponding identity of the $i_{th}$ user\\
         $N / d$ & Username/password pair\\
         $k_{s}$ & A set of SM2 public keys published regularly\\
         $\left(P_{S}^{i}, d_{S}^{i}\right)$ & The SM2 public/private key pairs published\\ & regularly in the system\\
         $K_{C}^{i}$ & User's SM4 key\\
         $\left(P_{C}^{i}, d_{C}^{i}\right)$ & Public/private key pair for signing and verifying\\
         $( r , s )$ & SM2 signature value\\
         $S M 2 E: \left\{N D, P_{S}^{i}\right\} \rightarrow CT$ & The process of generating ciphertext $CT$ by encrypting\\ & plaintext data $ND$ with SM2 public key $P_{S}^{i}$ \\
         $S M 2 D: \left\{C T, P_{S}^{i}\right\} \rightarrow ND$ & The process of decrypting $C T$ using $d_{S}^{i}$ to get $ND$\\
         $S M 4: \left\{N D, K_{C}\right\} \rightarrow CT$ & The process of generating data ciphertext $C T$ by \\ & encrypting plaintext data $N D$ with SM4 key $K_{C}$\\ 
         $S M 2 S: \left\{M r a d w, d_{S}^{i}\right\} \rightarrow(r, s)$ & The data to be sent is signed with SM2 algorithm\\
        \hline
    \end{tabular}
    \label{tab:tab1}
    \vspace{-2ex}
\end{table*}

\subsection{Key Generation.} Throughout the whole data sharing life cycle, we must verify its validity, integrity, and non-repudiation. We thus generate a pair of asymmetric keys for each user and sign their shared data. To satisfy the aforementioned conditions, AuthROS implements the production and verification of the elliptic curve public key by referencing the key pair generation and public key verification criteria introduced by the SM2 algorithm~\cite{ref9}.

Assuming that the private key is $d_{C}^{i}$, and AuthROS computes the public key $P_{C}^{i}=\left[d_{C}^{i} G\right]$ using the multiple points fast algorithm of multiple elliptic curves\cite{Fast_multiple_point}, where $G$ is the base point of the elliptic curve and its order is a prime number. In the meantime, AuthROS also employ the SM2 key derivation function to build the appropriate public/private key pair $\left(P_{S}^{i}, d_{S}^{i}\right)$ for key exchanges. Specific algorithms can be found in the standard for SM2 algorithms~\cite{ref9}.

\subsection{Key Allocation.} Data is often transferred in plaintext across the transmission connection in the current TCP/IP network transmission framework, making it available for malevolent users of a third party to intercept the data and conduct a series of assaults such as replay attacks and man-in-the-middle attacks. To maintain the security of data during transmission and storage on the blockchain, it is necessary to encrypt and authenticate the data using cryptographic technology. AuthROS implements the key distribution function to ensure secure storage of blockchain with SM4 and SM2. Among them, the SM4 key encrypts plaintext data and the SM2 public key confirms the digital signature.

Firstly, the user enters their own SM4 key $K_{C}^{i}$ and SM2 public key $P_{C}^{i}$. Then, the user selects a system public key $P_{S}^{i} \in k_{s}$ to encrypt the message $S M 2 E:\left\{\left(K_{C}^{i}, P_{C}^{i}\right), P_{S}^{i}\right\} \rightarrow C T$, where $d_{S}^{i}$ and $P_{S}^{i} \in k_{s}$ form a public/private key pair $\left(P_{S}^{i}, d_{S}^{i}\right)$, $k_{s}$ is a set of SM2 public keys regularly published by us. In addition, user should use their personal SM2 private key $d_{C}^{i}$ to sign the resulting ciphertext $C T, S M 2 S:\left\{C T, d_{S}^{i}\right\} \rightarrow(r, s)$. Then, adding an information frame after the ciphertext $C T$ to produce the final message $\left\{C T,(r, s), P_{S}^{i}\right.$, $Command$ $\}$, where $Command$ denotes the instructions of operations will be processed. The message is sent to the $C R$ via the TCP/IP protocol stack. $C R$ will decode the ciphertext with the SM2 private key $d_{S}^{i}$ corresponding to $P_{S}^{i}$, $S M 2 D:$ $\left\{C T, d_{S}^{i}\right\} \rightarrow\left(K_{C}^{i}, P_{C}{ }^{i}\right)$ after receiving the data packet. After decryption, $C R$ will utilize $P_{C}^{i}$ to validate the ciphertext's authenticity and integrity using $C T$, $S M 2 V:\left\{C T, P_{C}^{i},(r, s)\right\} \rightarrow$ $True$ $/$ $False$. After the integrity and accuracy checks, the keys $K_{C}^{i}$ and $P_{C}^{i}$ will be stored in the blockchain for next operations.

\subsubsection{Registration.} At this procedure, user $A U_{i}$ registers with AuthROS and uploads a key-value pair $\left(N_{i}, t_{i}\right)$ as the identity. Then, the $N_{i}$ supplied by the user and $t_{i}$, the SM4 key $K_{C}^{i}$ uploaded by the user during the key allocation, the SM2 public key $P_{C}^{i}$, and an unused account address $A d d r_{i}$ on the Ethereum network are used to generate a mapping $N_{i} \rightarrow\left(t_{i}, K_{C}^{i}, P_{C}^{i}, A d d r_{i}\right)$, In the preceding phase of key allocation, the user-uploaded SM4 key $K_{C}^{i}$ will now be assigned to $\sigma, \sigma=K_{C}^{i}, v$ and $v$ will remain empty, which means $(v=N u l l) \wedge(d=N u l l)$. The user's identification in the blockchain $E U_{i}\left(d, \sigma, v, A d d r_{i}\right)$ will be created.

\begin{figure}[ht]
\setlength{\abovecaptionskip}{0cm} 
\setlength{\belowcaptionskip}{-5cm}
    \centering
    \includegraphics[width=0.5\textwidth]{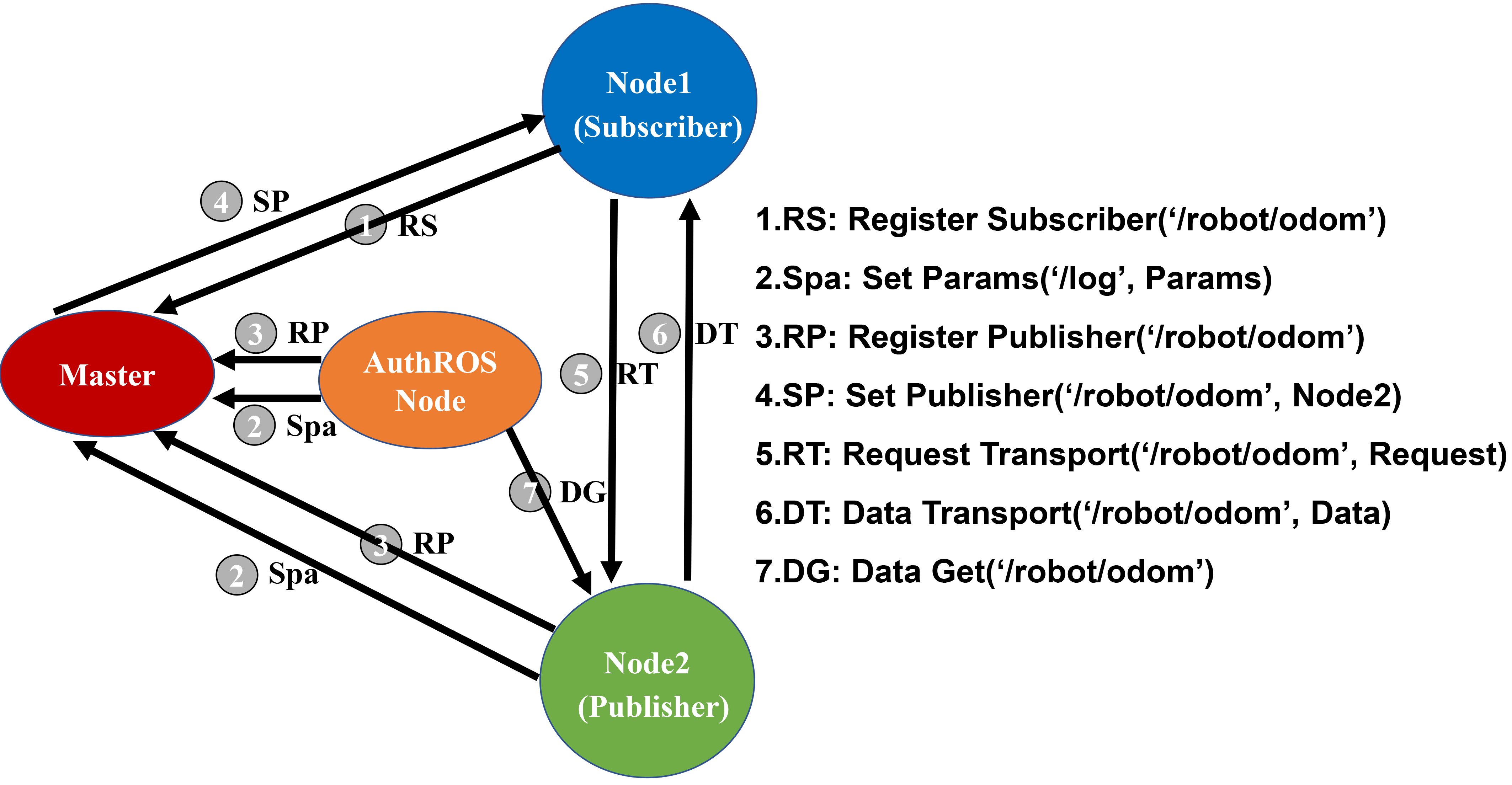}
    \caption{The workflow of ROS in local robots. Firstly, node1, node2, and AuthROS node should register their identity (subscriber or publisher) through ROS Master and set the topic ‘/robot/odom’ for message transmission. Then, the AuthROS node can seize data at any given time when the communication between node1 and node2 is going on. Then, extracted messages will go through a range of format conversion and encryption.}
    \vspace{-3ex}
    \label{fig:fig5}
\end{figure}

\subsection{Topic Set.} The core of AuthROS is a monitoring node. AuthROS is capable of monitoring the topics they want by inputting the names and message types of topics by users. Corresponding processing operations will be made for different message types. For example, for odometry-type messages, the AuthROS parses the respective messages after obtaining them from a topic named '1/robot/odom', and extracts useful information like three-axis velocity and angular velocity at a given time, etc. Fig. \ref{fig:fig5} shows how AuthROS obtain and parse the odometry-type messages.

\subsubsection{Encryption and Transfer.} We use the SM4 encryption method~\cite{ref10} to encrypt the $N D$ to get the ciphertext $C T$. $C T$ is packaged with some necessary information frames and sent to $C R$. The odometry-type data in ROS consists of information such as $L V_{x, y, z}, A V_{x, y, z}$. Before transmission, the data must be packaged into $\left\{L V_{x, y, z}, A V_{x, y, z}\right.$, $P o s e$, $T S$, $C o v$, $T y p e$ $\}_{ N D_{1}}$ and encrypted with SM4 for the first time $S M 4:\left\{N D_{1}, K_{C}^{i}\right\} \rightarrow C T_{1}$. The user signs the $C T_{1}$ with the SM2 private key $d_{C}^{i}$ to generate the signature value $S M 2 S:\left\{C T_{1}, d_{S}^{i}\right\} \rightarrow(r, s)$, encapsulates the information frame $\left\{C T_{1}\right.$, $T y p e$ $, t_{i}, T$, $C o m m a n d$,$\left.(r, s)\right\}_{N D_{2}}$ for the received ciphertext $C T_{1}$, re-encrypts the $N D_{2}, S M 4:\left\{N D_{2}, K_{C}\right\} \rightarrow C T_{2}$, encapsulates the information frame $\left\{C T_{2}, N_{i}\right\}_{N D_{3}}$ for the $C T_{2}$, and transmits it to the $C R$. After the identity check is successful, AuthROS will decrypt $C T_{2}$ to $C T_{1}$ and follow up the ciphertext $C T_{1}$ with the data digest interaction scheme according to $Command$. However, for non-general data types, such as image data, matrix and compression are necessary as pre-processing steps.

\section{Implementation and Evaluation}
This section introduces the hardware platform and smart contracts in experiments and then analyzes AuthROS of response time from different perspectives: Consensus Algorithms, Message Size, and Efficiency of the SM algorithm family. We evaluate the response time of data upload based on 4 ROS-Melodic robots with a ROS Master and a private Ethereum network on the host. For the same configuration of robots, we only evaluate the performance of a single robot.

\subsection{Hardware Platform Equipment}
The robots we used to equip with a Jetson Nano B01 (Quad-core ARM A57 64-bit @1.43Ghz 4GB LPDDR4-3200), a controller which has a built-in 9-axis IMU sensor, RPLIDAR A1 radar, and a Wi-Fi module that can provide up to 867mbps communication bandwidth and an RGB-D binocular camera. To realize the autonomous movement of the robot in the closed experimental environment, Visual Slam (Visual Simultaneous Localization and Mapping) and Lidar Slam (Lidar Simultaneous Localization and Mapping) are combined to build a complete map of the closed experimental space. ROS Melodic is set up in every robot, as long as the personal computer running ROS Master. Through the Wi-Fi module, each robot can connect to the ROS Master to achieve stable communication. The PC running ROS Master also connects to the server hosting the Ethereum network, thus can provide interaction between robots and Ethereum.

The controller and Jetson Nano are connected through a UART connection using software function calls provided by the controller's onboard C++ SDK. The controller collects the 9-axis IMU sensor and motor data. The RGB-D and RPLIDAR are connected to the Jetnano to capture images and collect lidar data. Motion commands are communicated between the Jetson Nano and controller to realize motion planning and control. The cloud server hosting Ethereum Network with a CPU (Intel Xeon (Ice Lake) Platinum 8369B @3.5GHz), memory (16GB DDR4 3200MHz), and disk (80GB ESSD).

\subsection{Ethereum Smart Contract Implementation}
The smart contracts developed in experiments are written in Solidity v7.6. The contracts allowed for identity registration, knowledge-upload, authority-grant, etc. All functions are listed as follows:

\textbf{Register(bytes).} This function registers an identity in the Ethereum network inputting a parameter of Bytes-type. The SM4 key is used for data encryption as a token. The robot owner converted the SM4 key to Bytes-type.

\textbf{Data Upload(bytes, bytes, bytes).} The robot owner uploads confidential data to the Ethereum network for immutable persistent storage by calling this method. This function accepts three parameters of Bytes-type, the first parameter is the data ciphertext of Bytes-type to be uploaded. The second parameter is a Bytes-type token (SM4 key) that indicates the identity of the data uploader. The third parameter is the timestamp of Bytes-type when this method was called.

\textbf{Authority Grant(address).} The robot that calls this method will append the token (SM4 key) that indicates its identity to the token list in the specified EOA account so that the specified account will have access to the function caller's data. This function accepts a parameter of address-type, which is the address of the EOA that access to the function caller's data. 

\textbf{Data Query(bytes, address).} The user can call this function to query the data being shared by the target EOA, when the user's token exists in the target EOA's token list. The first parameter ``bytes'' is a Bytes-type token that indicates the identity of the function caller, and the second parameter ``address'' indicates the target EOA address for the query.

\subsection{Evaluation Process}
\begin{figure*}[ht]
\setlength{\abovecaptionskip}{0cm} 
\setlength{\belowcaptionskip}{-5cm}
    \centering
    \includegraphics[width=0.8\textwidth]{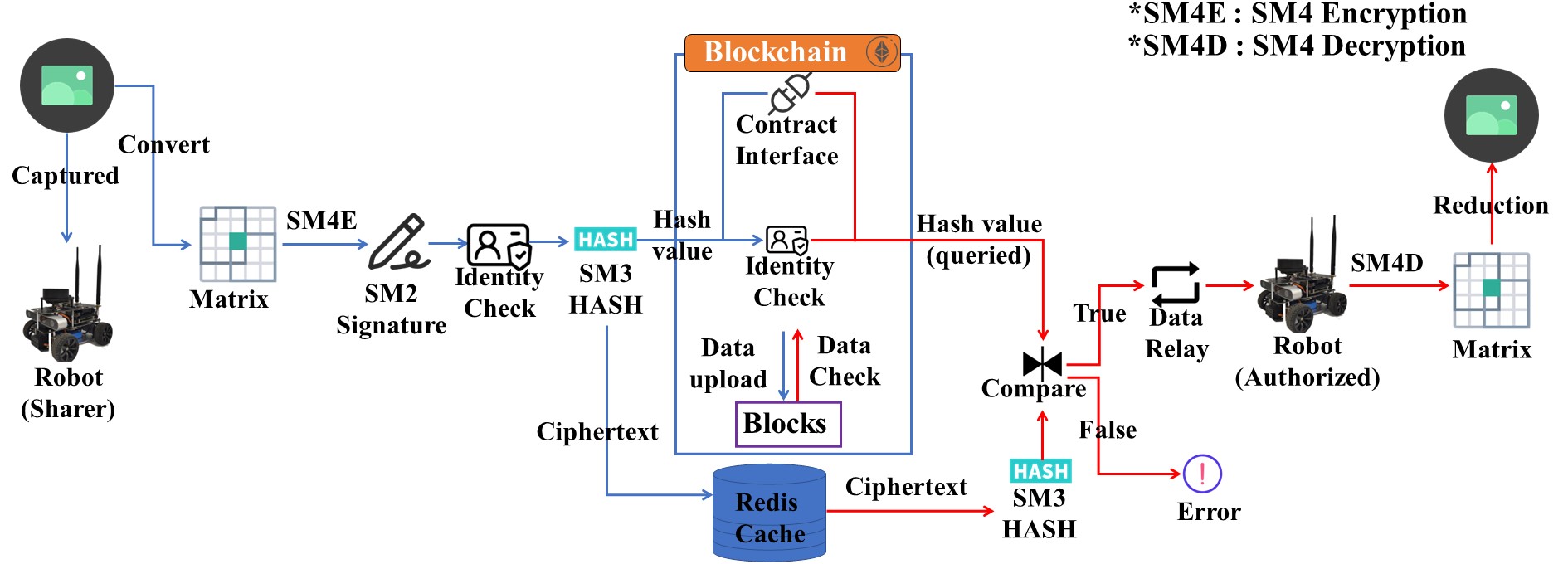}
    \caption{Process of image sharing. Images captured by robots are converted into a matrix. Then the matrix will be encrypted by the SM4 encryption algorithm, signed by the SM2 signature algorithm, and transmitted to the cloud server for persistence storage. The image is queried using the SM3 hash and ciphertext check.}
    \vspace{-2.5ex}
    \label{fig:fig7}
\end{figure*}

We take secure image sharing as an example in the AuthROS experiment. In the process of crime scene investigation and evidence collection, the images taken by different robots at the crime scene have strict requirements on security performance. User authorization is required for archiving and retrieval. We will carry out experiments against this, and the process of which is depicted in Fig. \ref{fig:fig7}.

Firstly, we start the robot and load the ROS master on the PC. After the initialization is complete, the robot will automatically connect to the ROS master according to the genesis block. We get the facial data we need from the topic '/robot/CompressedImage' in the form of a picture of 58 KB, and then use the OpenCV toolkit to matrix it. Next, the matrix will be converted into a character string and encrypted by SM4 algorithms. The ciphertext will be signed with the user’s SM2 public key and transmitted to the cache loaded in the server of AuthROS. At the same time, the SM3 cipher hash algorithm is used to generate the abstract of ciphertext, and the smart contract sends a transaction with the abstract to the Ethereum network with the help of the interface Data\_Upload.

Finally, the authorized robot owner can invoke the interface Data\_Check to check the ciphertext. The idea of homomorphic encryption is used for reference to design the process of data checking. The ciphertext within the cache is hashed by SM3, and the generated hash value is compared with the one queried from Ethereum. To ensure the immutability of data, if the two abstracts are the same, the Redis, an in-memory storage structure, will return ciphertext to the user, otherwise, return an error. At the same time, once a robot is authorized, it means that its owner has obtained the SM4 key of the data sharer. After the ciphertext is queried, the corresponding SM4 key can be used to decrypt the ciphertext, and the OpenCV can also be used to restore the image. The whole process can be seen in the video.

\subsection{Performance Evaluation}
\begin{figure}[t]
\setlength{\abovecaptionskip}{0cm} 
\setlength{\belowcaptionskip}{-5cm}
    \centering
    \includegraphics[width=0.5\textwidth]{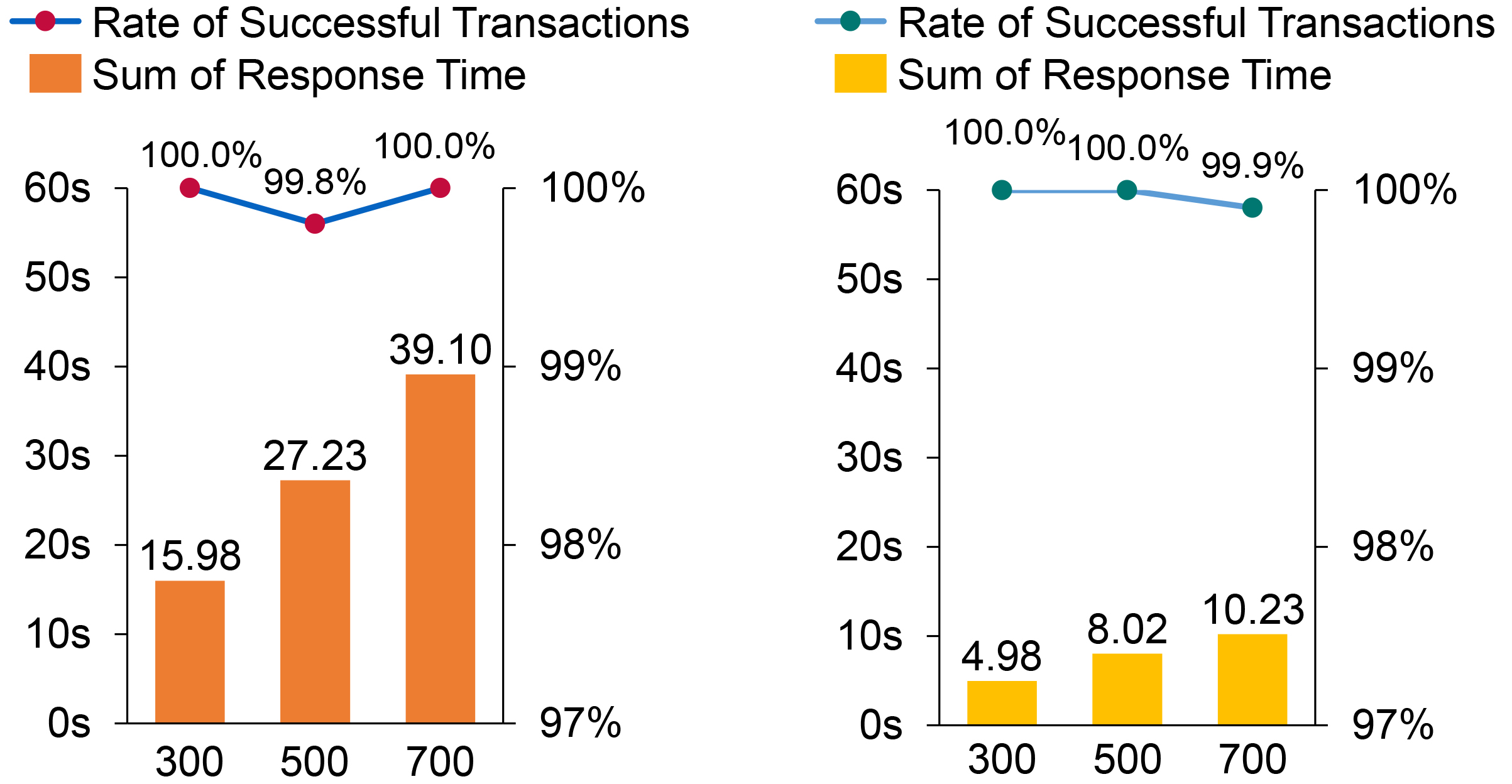}
    \caption{Left: total response time and success rate of PoW. Right: total response time and success rate of PoA.}
    \vspace{-2ex}
    \label{fig:fig11}
\end{figure}

\textbf{Consensus Algorithms.} As Fig. \ref{fig:fig11} shows, we set 300, 500, and 700 analog processes to send Data\_upload requests to the Ethereum network based on PoW and PoA consensus. We quantify the total response time and success rate of transactions under three different concurrencies. The networks share the same block difficulty, gaslimit, and message size, which are 0x4cccc8, 0xffffffff, and 1 KB respectively.

AuthROS maintains excellent interaction whichever consensus algorithm is used. In Fig. \ref{fig:fig11}, the success rate of transactions, which exceeds 99\% for both consensus algorithms, is comparable. However, in terms of response time, PoA consensus has obvious advantages over PoW consensus. This difference follows that PoA verifies transactions through preset nodes' voting. At the same time, PoW relies on the mining process to verify transactions. This process needs to consume many computing resources. We then conclude that PoA is more suitable for data transmission in robots.

\begin{figure}[t]
\setlength{\abovecaptionskip}{0cm} 
\setlength{\belowcaptionskip}{-5cm}
    \centering
    \includegraphics[width=0.45\textwidth]{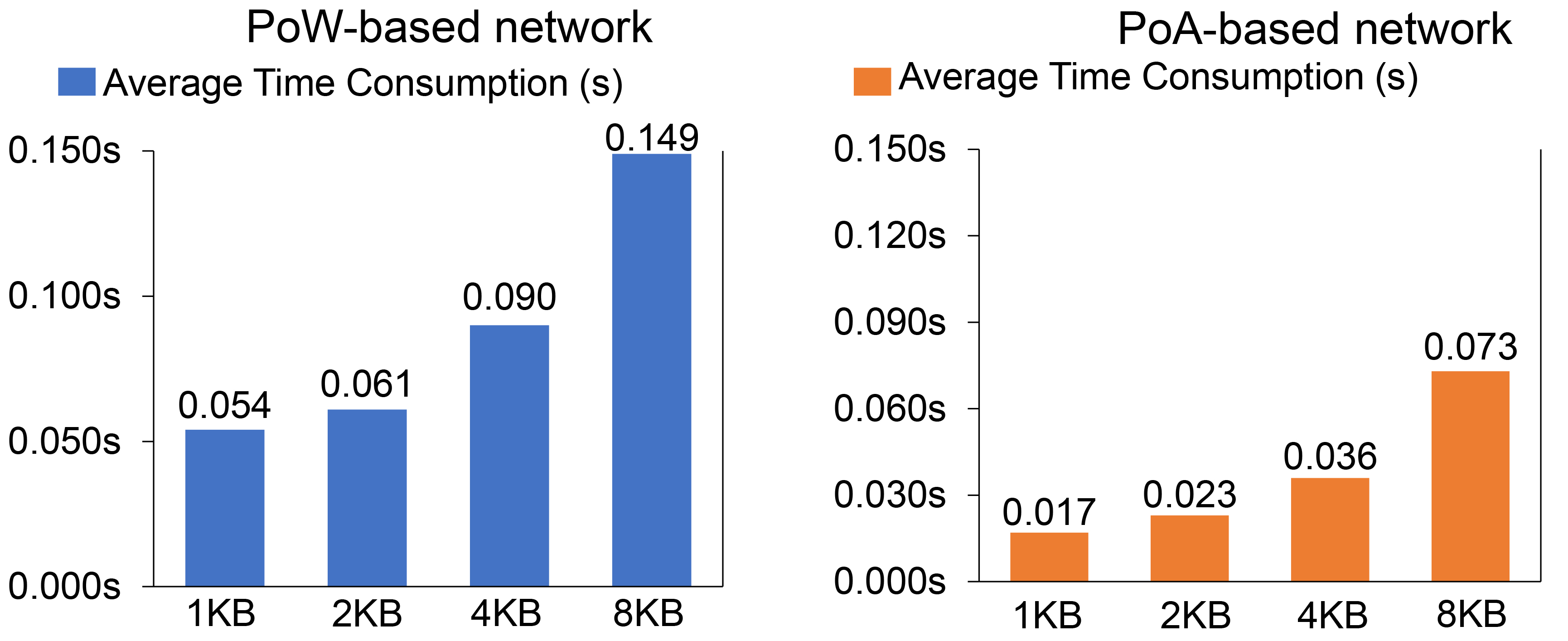}
    \caption{Left: average time consumption for different sizes of messages in seconds for a PoW-based network. Right: average time consumption for different sizes of messages in seconds for a PoA-based network.}
    \vspace{-2ex}
    \label{fig:fig12}
\end{figure}

\textbf{Messages Size.} When robots interact with the contract, the size passed to the contract method would have an impact on the response time. Therefore, we conduct related experiments to study the effect of message size on blockchain networks based on two different consensus algorithms. Message size is set to 4 values of 1KB, 2KB, 4KB, and 8KB. We call the Data\_upload interface 300 times through the simulated process and record the average time in Fig. \ref{fig:fig12}.

As Fig. \ref{fig:fig12} depicted, whether the Ethereum network is based on PoW or PoA consensus, the response time grows as message size increases. However, the PoA-based Ethereum network has a shorter overall average response time than the PoW-based Ethereum network. Furthermore, as message size rises, the average response time of the PoA-based Ethereum network grows more slowly.



\textbf{Efficiency and Stability of SM Algorithm Family.} In terms of data communication, AuthROS is equipped with key exchange based on SM2 to ensure the security of the SM4 key. Meanwhile, SM3 is used to generate the hash value of data in a big size. Thus, the efficiency and stability of SM algorithms have a huge impact on the availability and speed of AuthROS.



\begin{figure}[htbp]
\setlength{\abovecaptionskip}{-0.2cm} 
\setlength{\belowcaptionskip}{-5cm}
    \centering
    \includegraphics[width=0.5\textwidth]{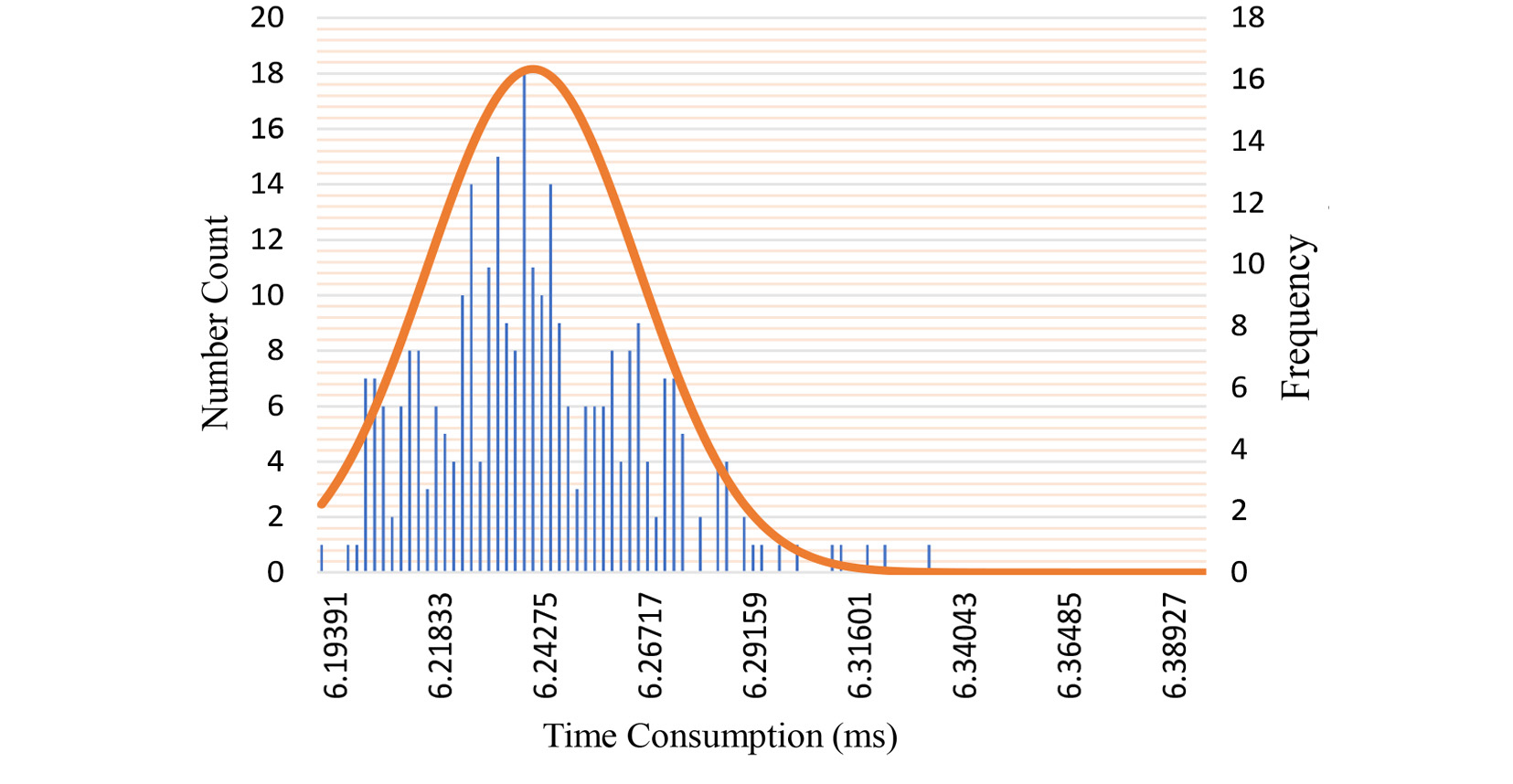}
    \caption{Time consumption of data digest generation each time (800 KB). The blue bar is for number count, and the orange line is for frequency.}
    \vspace{-1ex}
    \label{fig:fig17}
\end{figure}

\begin{figure*}[htbp!]  
	\centering
	\subfigure[data encryption]{
		\includegraphics[width=6in]{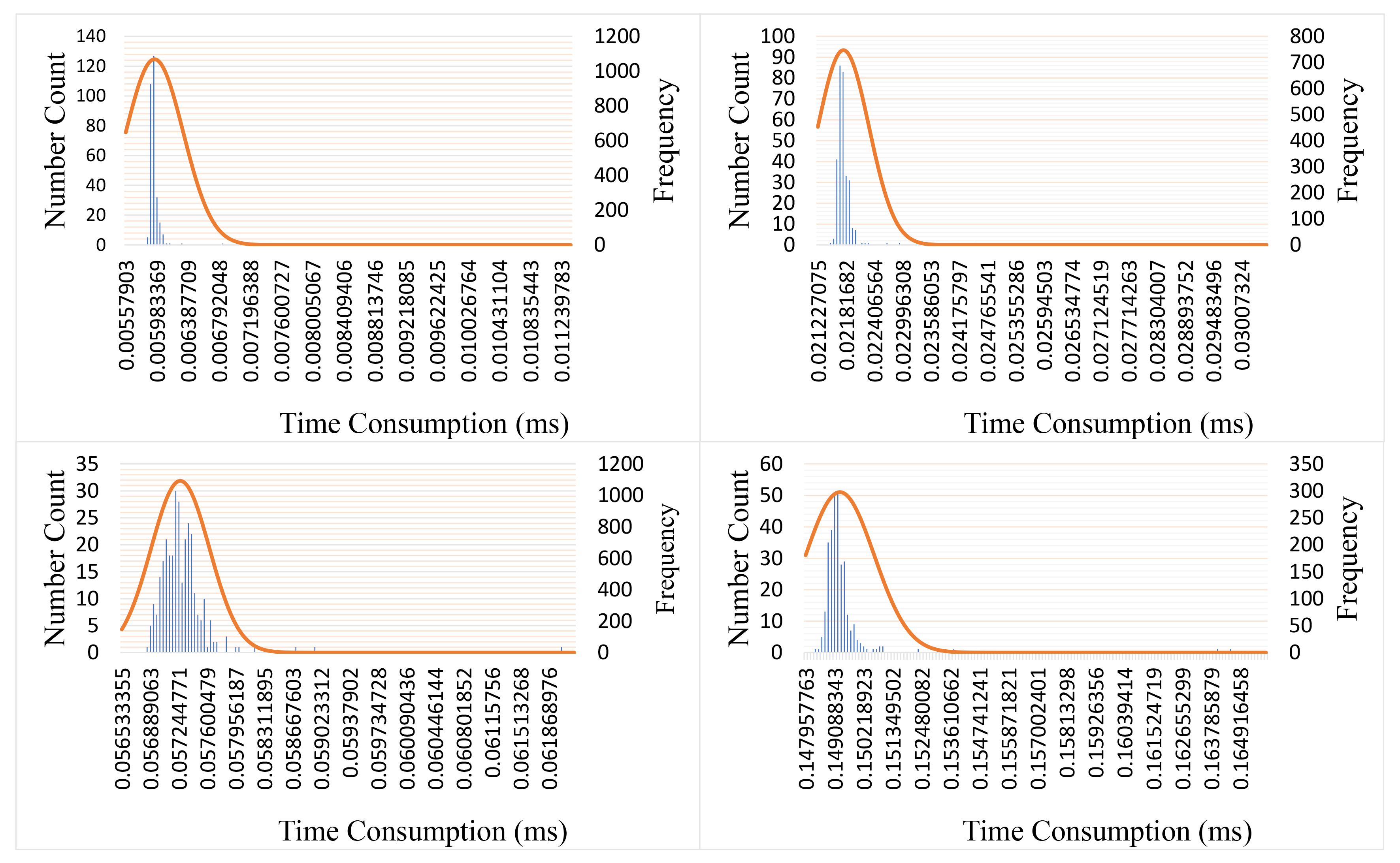}
	}
	
	\subfigure[data decryption]{
		\includegraphics[width=6in]{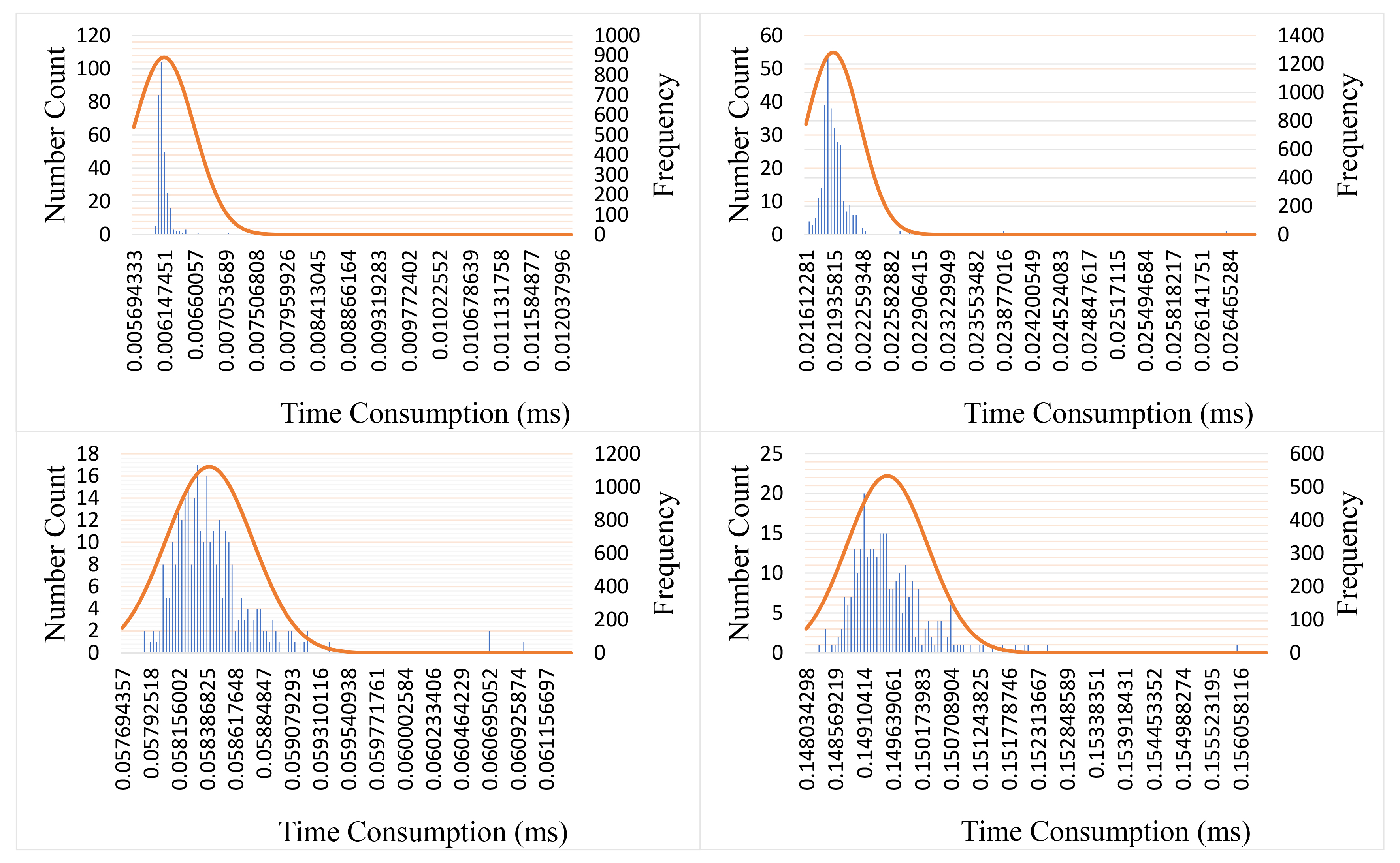}
	}
	\caption{Time consumption of 300 times data operations in different figure sizes (1KB, 2KB, 4KB, 8KB).  The blue bar is for number count, and the orange line is for frequency.}
	\label{fig:fig16}	
\end{figure*}

We conduct experiments on the encryption and decryption speed and stability of SM4 and SM3. We use plain-text data with sizes of 1KB, 2KB, 4KB, and 8KB as encrypted source data for SM4 encrypting and decrypting the data 300 times in Fig. \ref{fig:fig16} respectively, and recording the average time consumption. We also evaluate the speed and stability of SM3 using an 800 KB matrix as encrypted source data for digest generation 300 times in Fig. \ref{fig:fig17}.

\vspace{-1ex}

The encryption and decryption speeds of SM4 are close, and it is clear that the time consumption of decryption and encryption grows with the data sizes increase, which is due to the features of the symmetric encryption algorithm. However, it can be found in Fig. \ref{fig:fig16} that no matter the size of data, both decryption and encryption of SM4 have good stability where encryption time consumption concentrates in a certain range. It is the same with SM3. In Fig \ref{fig:fig17}, it is clear that the stability of the SM3 algorithm varies between 6.19ms and 6.34ms.

\section{Conclusion}
This paper proposes AuthROS, a novel data sharing framework for ROS, leveraging the Ethereum blockchain and SM algorithms. AuthROS is equipped with a key exchange mechanism and an authority granting mechanism. The key exchange mechanism guarantees the security of the SM4 key used for data encryption, and the authority granting mechanism ensures the trustworthiness of shared data and the controllability of information data. Through systematic experimental evaluation, the security and efficiency of AuthROS are verified. This work is also potential in some other fields as federated learning \cite{ref34, ref35, ref38}, cloud-edge cooperate robotics \cite{ref36, ref37}, etc.

\section*{Acknowledgments}
A preprint has previously been published~\cite{zhang2022authros}. We thank Songjing Tao, Shuang Wu, Ming Tang, Zeping Tang, and Zhixuan Liang for their early contributions to this work. Bernie Liu in the preprint is the same author as Boyi Liu in this paper. We thank reviewers for their helpful comments. This research is partially supported by Early Career Research Starting Fund of Hainan University under Grant RZ2200001265.

\end{document}